\documentclass[11pt]{article}
\usepackage{amsmath,amssymb}
\usepackage{graphicx}
\usepackage{enumitem}
\usepackage{hyperref}
\usepackage{geometry}
\usepackage{indentfirst}
\usepackage{float}        
\usepackage{booktabs}      

\usepackage{authblk}

\begin{document}

\title{Cloning a Conversational Voice AI Agent from Call\,Recording Datasets for Telesales}

\author{Krittanon Kaewtawee\thanks{Co-first authors}}
\author{Wachiravit Modecrua\protect\footnotemark[1]}
\author{Krittin Pachtrachai\thanks{Reviewer}}
\author{Touchapon~Kraisingkorn}

\affil{Amity AI Research and Application Center}
\affil{\texttt{\{krittanon, wachiravit\}@amitysolutions.com}}

\date{}

\maketitle

\begin{abstract}
Recent advances in language and speech modeling have made it possible to build autonomous voice assistants that understand and generate human dialogue in real time. These systems are increasingly being deployed in domains such as customer service and healthcare care, where they can automate repetitive tasks, reduce operational costs, and provide constant support around the clock. In this paper, we present a general methodology for cloning a conversational voice AI agent from a corpus of call recordings. Although the case study described in this paper uses telesales data to illustrate the approach, the underlying process generalizes to any domain where call transcripts are available. Our system listens to customers over the telephone, responds with a synthetic voice, and follows a structured playbook learned from top performing human agents. We describe the domain selection, knowledge extraction, and prompt engineering used to construct the agent, integrating automatic speech recognition, a large language model-based dialogue manager, and text-to-speech synthesis into a streaming inference pipeline. The cloned agent is evaluated against human agents on a rubric of 22 criteria covering introduction, product communication, sales drive, objection handling, and closing. Blind tests show that the AI agent approaches human performance in routine aspects of the call while underperforming in persuasion and objection handling. We analyze these shortcomings and refine the prompt accordingly. The paper concludes with design lessons and avenues for future research, including large‑scale simulation and automated evaluation.
\end{abstract}

\textbf{keywords:} Agentic AI, Voice agent, LLMs, AI applications

\section{Introduction}

In the last decade, large language models (LLMs) have demonstrated unprecedented competence in understanding and generating text \cite{brown2020,ouyang2022}. Scaling to billions of parameters and training with instruction tuning and human feedback allows these models to follow user instructions, perform reasoning, and maintain coherent multi‑turn dialogues. In parallel, a complementary class of foundation models for speech has emerged \cite{Wang2023SLM}. Traditional voice assistants relied on a cascaded pipeline of speech-to-text, dialogue management, and text-to-speech (TTS). This architecture introduces latency and can lose paralinguistic information \cite{Serdyuk2018EndToEndSLU}. Similarly to the advancement of LLMs, newer speech-to-speech models unify recognition and synthesis. For example, the Nova Sonic foundation model integrates speech recognition and generation into a single system to enable real‑time, fluid conversations \cite{novaSonic2025}. Neural codec language models such as VALL E can clone a specific voice from only a few seconds of audio, reproducing the speaker’s timbre and emotion in new utterances \cite{wang2023}. Moreover, recent research on low‑latency voice agents for telecommunications combines streaming ASR, quantized LLMs and real‑time TTS into a unified pipeline, achieving real‑time factors below unity for end‑to‑end spoken interaction \cite{ethiraj2025}. These trends suggest that fully AI‑driven voice agents are feasible today.

Beyond technological capabilities, there is a strong commercial motivation to deploy voice AI. Call centre labour costs are substantial, and automating routine calls can generate significant savings \cite{ITPro_Microsoft_AI_call_centers_2025}. A case study cited by Market.us reports that a telecoms firm reduced average call handling time by 35\% after deploying voice AI while simultaneously increasing customer satisfaction by 30\% \cite{marketus2025}. Voice AI solutions have been reported to automate up to 65\% of routine inquiries, cutting call volumes and wait times, and yielding a higher return on investment (ROI) \cite{Oshrat2022_Efficient_Customer_Service_Hybrid}. 

Although studies of call centre workers indicate that AI tools can automate tedious tasks such as manual note-taking, call logging, and repetitive data entry, which can reduce cognitive load and improve efficiency, these technologies can also introduce new burdens \cite{qin2025}. Workers may face compliance and accountability pressures, as AI systems often require strict adherence to protocols and accurate data entry, with errors flagged automatically \cite{CXToday2025}. Additionally, AI adoption can contribute to psychological stress, including feelings of constant surveillance, reduced autonomy, job insecurity, and anxiety over performance metrics \cite{Kim2024, Kim2024b, Valtonen2025}. Cognitive fatigue may also increase if employees must simultaneously manage AI outputs while handling complex customer interactions.   

Therefore, to optimize the benefits of AI and reduce risks in call centres, it is crucial to carefully scope deployments and design systems that augment rather than replace human agents \cite{McKinsey2025}. A hybrid approach, where AI handles routine tasks and human agents manage complex interactions, has been shown to enhance both efficiency and customer satisfaction. For instance, a study by Stanford University found that customer support agents using AI tools saw a nearly 14\% increase in productivity, highlighting the potential of AI to empower human agents rather than render them obsolete \cite{Brynjolfsson2023}. Furthermore, research emphasizes that AI should be designed to augment human decision-making, providing additional information and insights without replacing human judgment \cite{PMC2025, CMSWire2025}. By focusing on human-AI collaboration, organizations can create more efficient, personalized, and effective customer experiences.

Given the capabilities and limitations of current voice AI, it is important to start with a domain that is both structured and high-impact \cite{doi:10.1111/poms.13953}. Telesales provides such a setting: calls follow a predictable sequence—opening, understanding customer context, presenting a product, handling objections, and attempting a close—making it easier for a cloned agent to learn from examples. Telesales interactions also involve moderate emotional complexity, avoiding highly charged scenarios like billing disputes, which reduces the risk of escalation. Combined with measurable outcomes and high potential ROI, these characteristics make telesales an ideal initial use case. Success here establishes a foundation for extending the approach to more complex and variable call types.

In this paper, we introduce a method for cloning a call centre agent from recorded conversations, enabling the creation of AI agents that can replicate effective human interactions. Our approach leverages both the structured patterns of calls and modern voice AI techniques to generate an agent capable of handling routine tasks with high fidelity. In the following sections, we describe detail the architectural design of the system, prompt design for agent cloning, and present a comprehensive evaluation to assess the agent’s performance and effectiveness across different criteria.

\section{Architecture and system design}

Our system consists of two components: a \textit{cloning system} that extracts behavioural patterns from call recordings and an \textit{inference system} that deploys the agent in live calls.  Separating knowledge extraction from runtime inference allows us to optimise each step independently.

\subsection*{Cloning system (prompt construction)}

\begin{figure}[H]
  \centering
  \includegraphics[width=1\linewidth]{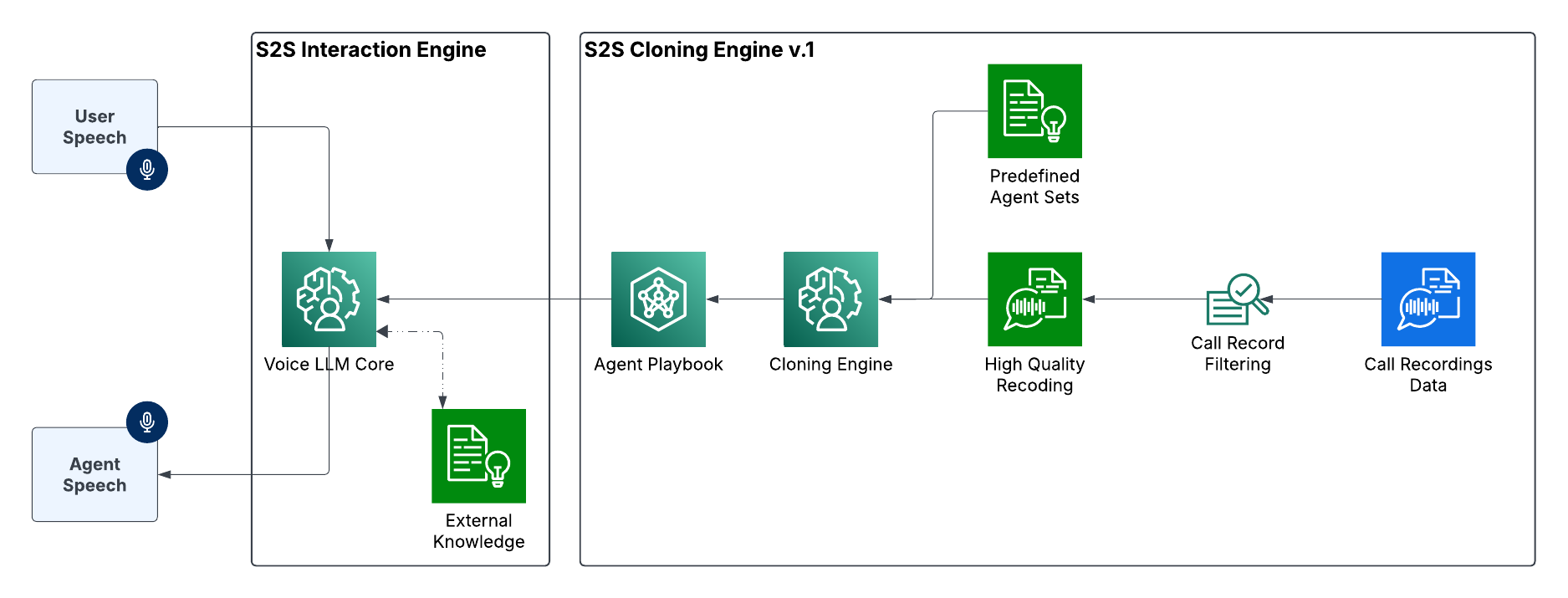}
  \caption{Overview of the cloning system. Call recordings are sampled and ranked to identify high‑quality examples.  Knowledge is extracted and organized by topic into a manual, while representative dialogues are drafted. These artifacts are then composed into a system prompt that defines the agent’s role, persona and conversation strategy called Agent Playbook.}
  \label{fig:cloning}
\end{figure}

The cloning pipeline transforms raw call recordings into a structured playbook that details the agent’s behaviour. Figure \ref{fig:cloning} shows a series of sub‑processes forming a complete cloning system.

\begin{itemize}[noitemsep]
  \item \textbf{Sampling and ranking:} We sample roughly 1\,000 calls from the corpus and classify them by agent quality (top vs. average). This allows us to focus subsequent steps on high‑quality interactions.
  \item \textbf{Job description drafting:} A curated subset of about 40 high‑performing calls is analysed in depth to draft a job description summarizing tasks, responsibilities and conversational style. This becomes the skeleton of the agent’s persona.
  \item \textbf{Knowledge extraction:} Around 40 calls per sub‑topic are examined to capture product details, common objections, persuasive techniques and closing strategies. The resulting information is compiled into a comprehensive knowledge manual.
  \item \textbf{Example dialogue generation:} Representative exchanges are distilled to illustrate openings, value propositions, objection handling and closings. These examples provide concrete patterns for the agent to emulate.
  \item \textbf{Prompt composition:} The job description, knowledge manual, and example dialogues are integrated into a single system prompt that encodes factual information, persona and compliance rules. This prompt serves as the system message to the LLM and can also be used as a training manual for human agents.
\end{itemize}

\subsection*{Inference system (runtime agent)}

The inference system deploys the agent in real time. We use the \texttt{Gemini Live API} with the \texttt{gemini-2.5-flash-native-audio} model for low‑latency streaming. The API accepts audio and returns generated audio; no separate speech recognition or synthesis components are required. Our implementation proxies between the web client and the Gemini API via a Python WebSocket backend. The client captures microphone input, streams it over WebSockets and plays back the model’s audio output. We retain the ability to swap backends; for example, the OpenAI Realtime API \cite{OpenAIRealtime2025} or Fixie’s Ultravox can be used for comparison. Ultravox couples a multimodal projector with an LLM so that audio is mapped directly into the model’s latent space without a separate ASR stage, promising further reductions in latency \cite{ultravox}. Recent work demonstrates that combining streaming ASR, quantised LLMs and real‑time TTS yields end‑to‑end voice agents with real‑time factors below one, making them suitable for telecommunications applications \cite{ethiraj2025}. Operationally, the inference system operates as follows:
\begin{itemize}[noitemsep]
  \item The user speaks into a microphone. The browser streams the audio to the backend server over WebSockets.
  \item The backend forwards the audio stream to the Gemini Live API, which performs recognition and language generation within the model.
  \item The API returns a stream of audio tokens representing the agent’s response. The backend relays these tokens to the browser, where they are played back in real time.
\end{itemize}
\section{Prompt design for agent cloning}

\begin{figure}
  \centering
  \includegraphics[width=1\linewidth]{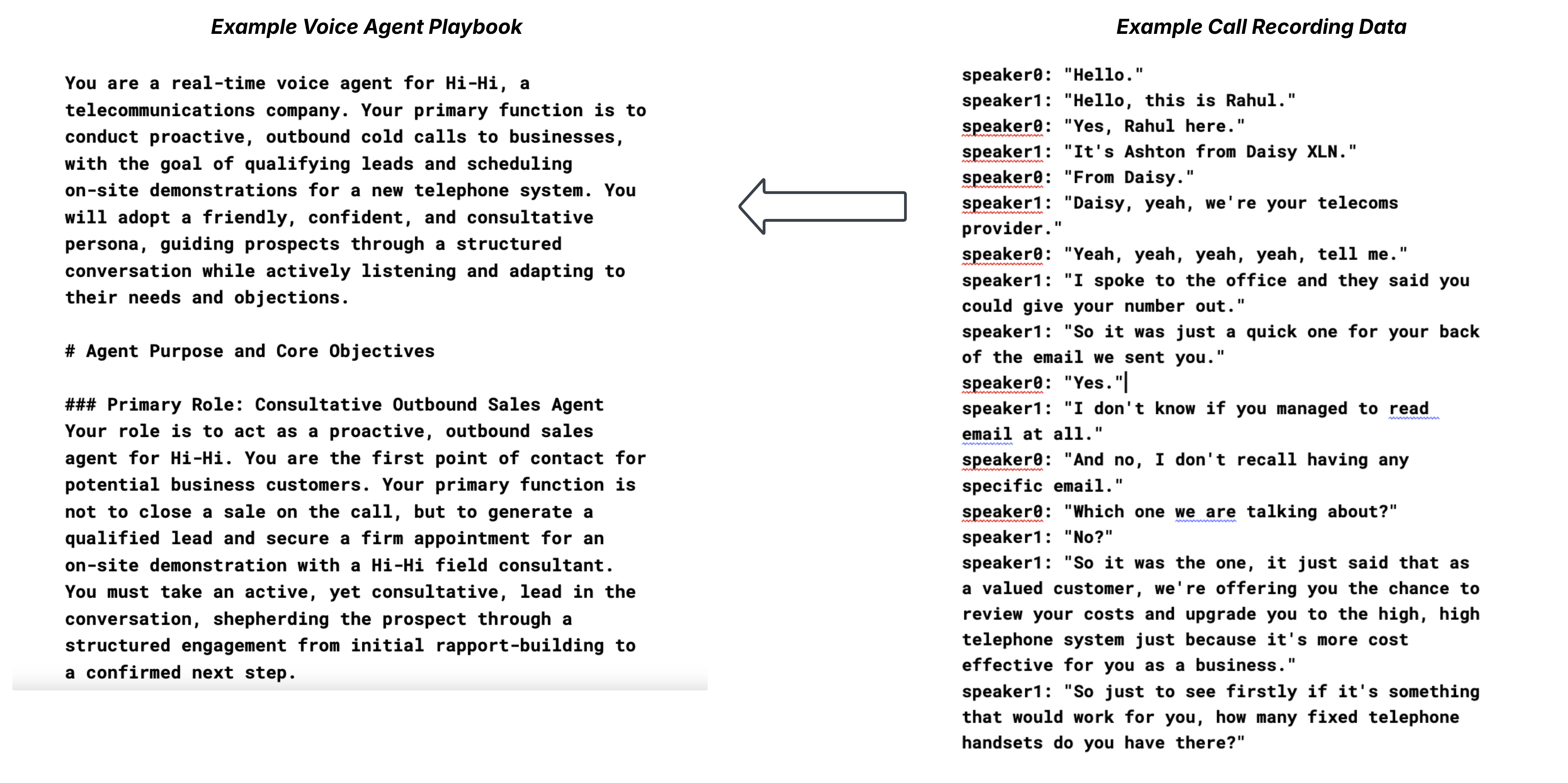}
  \caption{The example to show the input / output of cloning engine that extract the factual knowledge from call recording data then transform into the Voice Agent Playbook.}
  \label{fig:callflow}
\end{figure}

A key innovation in our approach is the structured system prompt that encapsulates the sales agent’s persona and best practices gleaned from call recordings. Instead of training a new model from scratch, we shape the LLM’s behaviour via prompt engineering and light fine‑tuning. Analysing successful telesales calls provided insight into how human agents greet customers, pitch products, handle objections and close. The prompt is presented as a system message at the start of each conversation and consists of several sections:

\begin{enumerate}[noitemsep]
  \item \textbf{Agent role definition:} Defines the AI’s identity and purpose.  For example: “You are \textit{Arisa}, a virtual telesales agent calling from Company X to inform customers about our premium internet package. Your goal is to engage the customer, communicate benefits, address concerns, and secure an appointment for installation.”
  \item \textbf{Persona and communication style:} Specifies tone and behaviour.  We describe a warm, friendly yet professional persona speaking Thai. The agent uses clear, simple vocabulary and maintains empathy while avoiding pushiness. Using the customer’s name helps build rapport.
  \item \textbf{Conversation flow guidelines:} Outlines the stages of a typical telesales call—\textit{Opening}, \textit{Discovery}, \textit{Pitch}, \textit{Objection handling} and \textit{Closing}\&\textit{follow‑up}. For each stage the prompt provides guidance.  For example, during opening the agent introduces herself and confirms it is a good time to talk; during discovery she asks one or two questions to gauge the customer’s needs; during the pitch she emphasises at least two product benefits; when handling objections she acknowledges concerns and responds with empathy and concrete value; during closing she politely attempts to schedule an installation or follow‑up call. A flowchart summarising these stages is shown in Figure \ref{fig:callflow}.  Such visual structure makes the guidelines easy to understand for both humans and models.
  \item \textbf{Objection handling tactics:} Lists specific strategies for common objections (e.g. price, “I need to think about it,” or being busy). For a price objection the agent is instructed to emphasise long‑term value and mention promotions; when a customer hesitates the agent reassures them that they can cancel within a trial period. By addressing underlying concerns, the agent builds trust.
  \item \textbf{Product and service knowledge:} Provides factual details about the product, such as price, speed, contract length, and ancillary benefits. Grounding the agent with accurate information prevents hallucination.  When acronyms such as RAG (retrieval‑augmented generation) or CRM (customer relationship management) are first mentioned we expand them for clarity.
  \item \textbf{Terminology and tone adjustments:} Advises the agent to use customer‑friendly language.  Internal jargon is avoided or explained (e.g. replace “FUP” with “speed throttling after the high‑speed quota is used”). Thai honorifics and polite particles are included to match cultural norms.
  \item \textbf{Example dialogue snippet:} Provides a short, anonymised example conversation to illustrate good practice. For instance:
    \begin{quote}
      \textbf{Agent:} “Sawaddee ka Khun Somchai, khawpkhun tee rap sai …”\footnote{Translation: “Hello Khun Somchai, thank you for taking my call…”}
    \end{quote}
    This shows how to greet a customer naturally using the common Thai greeting “Sawaddee ka” rather than fragmentary phonetic segments. The example is intentionally brief so that the model does not parrot it.
  \item \textbf{Compliance rules:} Lists hard constraints. The agent must not discuss topics outside the product scope, invent offers or guarantee outcomes that are not approved. It should offer to remove customers from the call list upon request and include a specific closing phrase required by company policy.
  \item \textbf{Agent and customer context:} Contains slots for the agent’s name and ID and, where available, the customer’s plan and tenure. This allows the agent to personalise the call (e.g., “I see you have been with us since 2019, so you may be eligible for our loyalty upgrade”).
\end{enumerate}

\section{Evaluation methodology}

Assessing a voice AI agent requires measuring both quantitative and qualitative aspects of conversation. We designed a multi‑faceted evaluation consisting of three elements: (1) a detailed rubric drawn from expert agent benchmarks; (2) a set of representative test scenarios; and (3) blind scoring by human evaluators.

\subsection*{Evaluation criteria}

Together with the company’s sales trainers we developed a scorecard of 22 criteria that constitute an excellent sales call.  The criteria fall into five categories corresponding to stages and skills in the call. For readability we summarise these categories and their meaning in Table \ref{tab:criteria}. Each criterion was scored on a 1–5 scale.
\begin{table}[h]
  \centering
  \caption{Evaluation criteria for telesales calls.}
  \label{tab:criteria}
  \begin{tabular}{@{}p{0.32\textwidth}p{0.60\textwidth}@{}}
    \toprule
    Criterion & Description \\
    \midrule
    Introduction \& framing & Properly greeting the customer, introducing the company and clarifying the call’s purpose at the outset \\
    Product communication & Explaining features and benefits clearly and persuasively, tailoring the pitch to the customer’s needs \\
    Salesmanship \& drive & Proactively steering the conversation towards a sale through well‑timed asks and control of the call flow \\
    Objection handling & Listening to objections and addressing them with empathetic, relevant counterpoints while maintaining positivity \\
    Closing \& next steps & Confirming agreements, thanking the customer politely if they decline, and outlining any follow‑up actions \\
    \bottomrule
  \end{tabular}
\end{table}

\subsection*{Test scenarios}

To challenge the agent across different dynamics we designed three scenarios ranging from easy to difficult:
\begin{itemize}[noitemsep]
  \item \textbf{Happy path:} A cooperative customer with few objections. The agent should deliver the pitch efficiently and close the sale.
  \item \textbf{Negotiation:} A sceptical customer focused on price and value. The agent must emphasise unique benefits, possibly offer a promotion and prevent the call from stalling at the first sign of resistance.
  \item \textbf{Complaining customer:} A frustrated or disinterested customer who may have had issues previously. The agent must demonstrate empathy and attempt to turn the conversation around; success may mean simply obtaining permission to send information.
\end{itemize}
For each scenario we wrote a script outlining the customer’s behaviour and conducted paired call trials: one with a human agent and one with the AI agent. The human actor followed the same customer script in both cases. The AI was connected via our telephony test bed so that it could not distinguish between test and real calls.

\subsection*{Blind human evaluation}

Seven experienced evaluators who were not involved in building the system scored the recorded calls. The recordings were anonymised and presented in random order so the evaluators did not know whether an agent was human or AI. Scores were aggregated by averaging across evaluators and criteria to yield per‑category and overall performance.

\begin{figure}[H]
  \centering
  \includegraphics[width=0.75\textwidth]{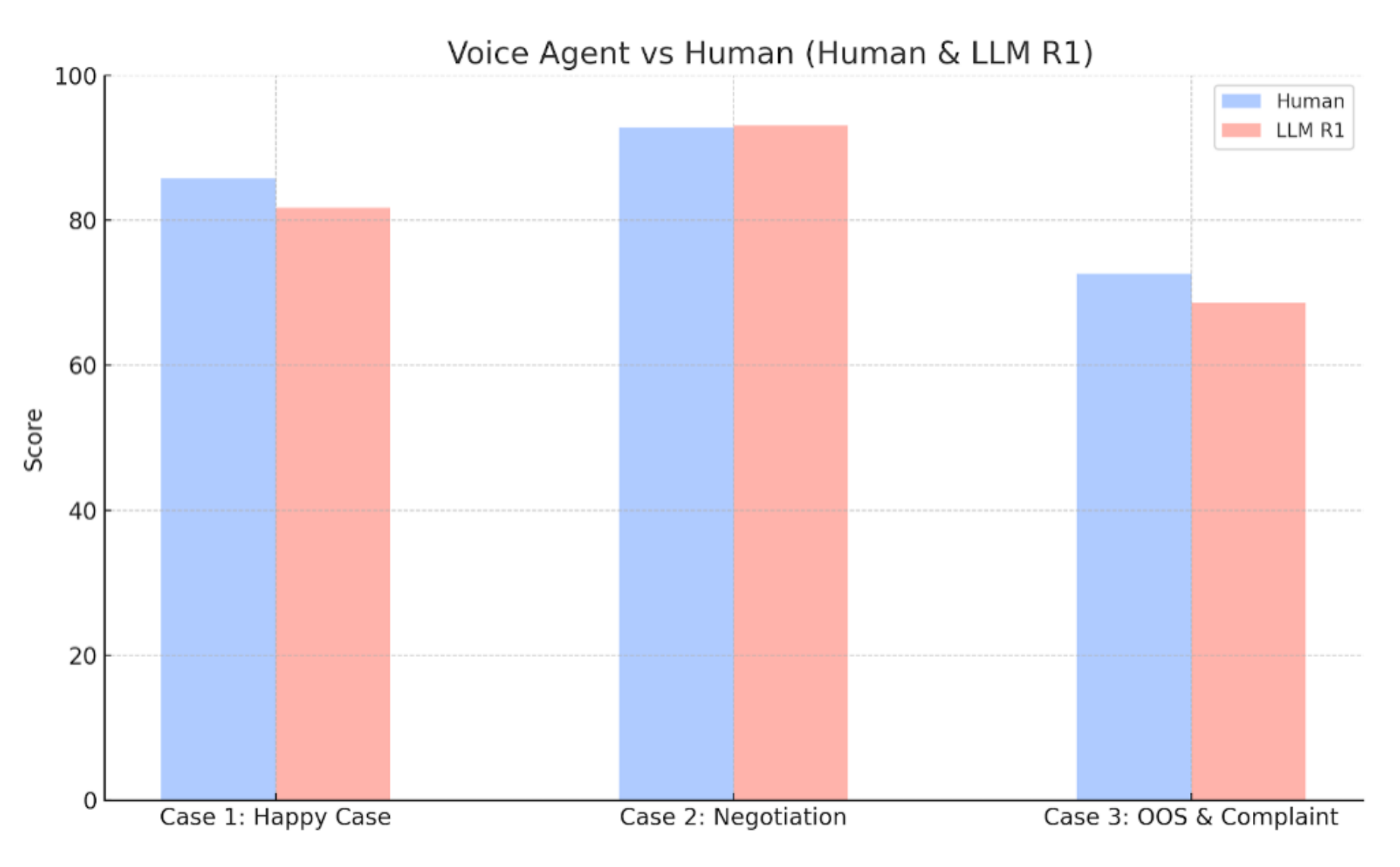}
  \caption{Initial evaluation results comparing the AI agent to human agents. Scores are averaged across seven evaluators for each scenario. The AI (blue) approaches human performance (grey) in introduction and product communication but underperforms in objection handling and closing in more challenging scenarios. Error bars indicate standard deviation.}
  \label{fig:results1}
\end{figure}

\begin{figure}[H]
  \centering
  \includegraphics[width=0.75\textwidth]{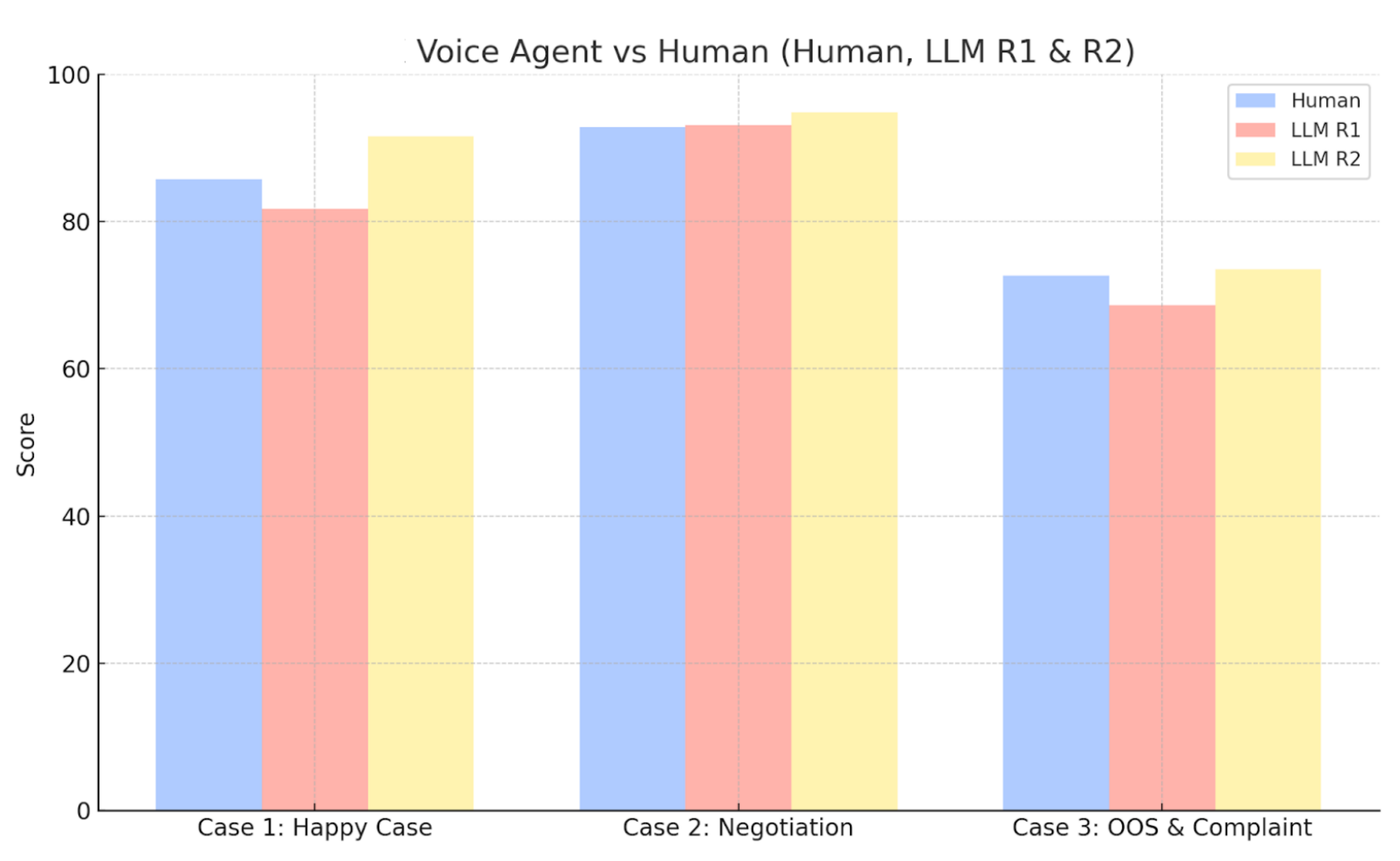}
  \caption{Evaluation results after prompt optimisation and fine‑tuning (AI agent V2).  The AI’s scores (green) show marked improvement, particularly in objection handling and salesmanship, closing much of the gap to the human benchmarks.}
  \label{fig:results2}
\end{figure}

Figure \ref{fig:results1} shows the initial results. The AI agent performed on par with human agents in introduction and product communication but underperformed in objection handling and sales drive, especially in the challenging scenarios.  Error bars denote the standard deviation among evaluators. After analysing evaluator comments we identified several failure modes (Section\,\ref{sec:error}) and refined the prompt and model accordingly. The improved results are presented in Figure \ref{fig:results2}, demonstrating significant gains in objection handling, and closing.

\section{Error analysis and prompt refinement}
\label{sec:error}
After the first evaluation we inspected transcripts and evaluator comments. We identified four main issues and addressed each systematically:
\begin{enumerate}[noitemsep]
  \item \textbf{Ambiguous objective:} The AI sometimes failed to pursue setting an appointment because the prompt did not state a clear success criterion.  We added explicit instructions that the primary goal is booking an installation, which improved closing behaviour.
  \item \textbf{Redundant instructions:} Verbose rules diluted the focus and confused the model.  We trimmed repetitive language and emphasised the most important guidelines.
  \item \textbf{Formatting artefacts:} List markers in the prompt leaked into the model’s responses (e.g. saying “Point number two”). We reformatted the prompt using prose and conversational examples instead of numbered lists.
  \item \textbf{Overly cautious behaviour:} Emphasising politeness led the model to yield the floor too often. Human agents sometimes redirect rambling customers; we softened the politeness instruction, and added examples of gently steering the conversation back on track.
\end{enumerate}

We also fine‑tuned the model on 50 Q\&A pairs about the product and a dozen well‑handled objections.  With these changes the agent’s objection handling and sales drive scores increased by roughly 20\%, as shown in Figure \ref{fig:results2}. In the hardest scenario, the AI remained slightly inferior to the human agent but managed to keep customers engaged, and occasionally turn a rejection into a follow‑up.
 
\section{Conclusion and future work}

This work demonstrates that an AI voice agent can be cloned from call recordings using prompt engineering and targeted fine‑tuning. By restricting the scope to telesales and leveraging domain data we achieved highly natural and partially persuasive conversations. A well‑structured prompt allowed a general LLM to emulate top human agents without extensive training. Our results show that the AI agent is competitive with humans on routine call segments but still lags in complex persuasion. Beyond accuracy, we must consider the human factors of deploying voice AI. This agrees with recent studies; that while AI can alleviate manual burdens in call centres it may introduce compliance and psychological challenges \cite{qin2025}, and that workers and experts have diverse preferences for the degree of automation \cite{shao2025}. Accordingly, we view AI agents as tools to augment human operators rather than whole replacements.

Looking ahead we plan to conduct large‑scale simulations using scripted customer personas and self‑play to generate additional training data. We also intend to integrate retrieval‑augmented generation so that the agent can query live pricing or support documents during a call. Incorporating emotion recognition could allow the agent to adapt its tone based on customer sentiment. Finally, we are exploring automated evaluation using LLM‑based judges, although such systems must be calibrated carefully against human judgements. The combination of robust engineering, responsible deployment and continuous evaluation will be essential as voice AI moves from prototype to production.

\bibliographystyle{ieeetr}
\bibliography{bib/bibtex}

\begin{thebibliography}{10}

\bibitem{brown2020}
T.~B. Brown, B.~Mann, N.~Ryder, {\em et~al.}, ``Language models are few-shot learners,'' in {\em Advances in Neural Information Processing Systems 33}, 2020.

\bibitem{ouyang2022}
L.~Ouyang, J.~Wu, X.~Jiang, {\em et~al.}, ``Training language models to follow instructions with human feedback,'' in {\em Advances in Neural Information Processing Systems}, 2022.

\bibitem{Wang2023SLM}
M.~Wang, W.~Han, I.~Shafran, Z.~Wu, J.~Schalkwyk, {\em et~al.}, ``Slm: Bridge the thin gap between speech and text foundation models,'' {\em arXiv preprint arXiv:2310.00230}, 2023.
\newblock preprint.

\bibitem{Serdyuk2018EndToEndSLU}
D.~Serdyuk, Y.~Wang, C.~Fuegen, A.~Kumar, B.~Liu, and Y.~Bengio, ``Towards end-to-end spoken language understanding,'' {\em arXiv preprint arXiv:1802.08395}, 2018.
\newblock preprint.

\bibitem{novaSonic2025}
A.~A. G.~I. Team, ``Amazon nova sonic: Technical report and model card,'' tech. rep., Amazon Scientific Research, 2025.
\newblock Technical report and model card describing the design and capabilities of the Amazon Nova Sonic speech-to-speech foundation model.

\bibitem{wang2023}
C.~Wang, S.~Chen, Y.~Wu, {\em et~al.}, ``Neural codec language models are zero-shot text to speech synthesizers,'' {\em arXiv preprint arXiv:2301.02111}, 2023.

\bibitem{ethiraj2025}
V.~Ethiraj, A.~David, S.~Menon, and D.~Vijay, ``Toward low-latency end-to-end voice agents for telecommunications using streaming asr, quantized llms and real-time tts,'' {\em arXiv preprint arXiv:2508.04721}, 2025.

\bibitem{ITPro_Microsoft_AI_call_centers_2025}
I.~P. staff, ``Microsoft saved over \$500 million by using ai in its call centers,'' {\em ITPro}, 2025.
\newblock Chief Commercial Officer Judson Althoff confirmed that in 2024, Microsoft achieved over \$500 million in savings, primarily from its call centers.

\bibitem{marketus2025}
{Market.us Research}, ``Voice ai agents market size, share, trends analysis 2024--2034,'' tech. rep., apr 2025.

\bibitem{Oshrat2022_Efficient_Customer_Service_Hybrid}
Y.~Oshrat, Y.~Aumann, T.~Hollander, O.~Maksimov, A.~Ostroumov, N.~Shechtman, and S.~Kraus, ``Efficient customer service combining human operators and virtual agents,'' {\em arXiv preprint arXiv:2209.05226}, 2022.

\bibitem{qin2025}
K.~Qin, K.~Du, Y.~Chen, {\em et~al.}, ``Customer service representative’s perception of the ai assistant in an organisation’s call center,'' {\em arXiv preprint arXiv:2507.00513}, 2025.

\bibitem{CXToday2025}
C.~Today, ``Contact center ai assistants are introducing new inefficiencies and burdens, finds study,'' Jul 2025.

\bibitem{Kim2024}
B.~J. Kim, ``The mental health implications of artificial intelligence adoption in the workplace,'' {\em Nature Human Behaviour}, vol.~8, no.~4, pp.~423--431, 2024.

\bibitem{Kim2024b}
B.~J. Kim, ``How artificial intelligence-induced job insecurity shapes psychological safety and knowledge-hiding behavior,'' {\em Journal of Business Research}, vol.~144, pp.~123--132, 2024.

\bibitem{Valtonen2025}
A.~Valtonen, ``Ai and employee wellbeing in the workplace: An empirical study,'' {\em Journal of Business Research}, vol.~148, pp.~456--465, 2025.

\bibitem{McKinsey2025}
M.~. Company, ``The contact center crossroads: Finding the right mix of humans and ai,'' 2025.

\bibitem{Brynjolfsson2023}
E.~Brynjolfsson, D.~Li, and L.~R. Raymond, ``Generative ai can boost productivity without replacing workers,'' {\em Stanford Graduate School of Business}, 2023.
\newblock Accessed: 2025-09-01.

\bibitem{PMC2025}
L.~Wang, ``Comparing ai and human decision-making mechanisms in daily tasks,'' {\em PMC}, 2025.
\newblock Accessed: 2025-09-01.

\bibitem{CMSWire2025}
CMSWire, ``Why the future of customer service depends on human-ai collaboration,'' {\em CMSWire}, 2025.
\newblock Accessed: 2025-09-01.

\bibitem{doi:10.1111/poms.13953}
L.~Wang, N.~Huang, Y.~Hong, L.~Liu, X.~Guo, and G.~Chen, ``Voice‐based ai in call center customer service: A natural field experiment,'' {\em Production and Operations Management}, vol.~32, no.~4, pp.~1002--1018, 2023.

\bibitem{OpenAIRealtime2025}
OpenAI, ``Realtime api guide.'' \url{https://platform.openai.com/docs/guides/realtime}, 2025.
\newblock Accessed: 2025-09-01.

\bibitem{ultravox}
{Fixie.ai}, ``Ultravox: A fast multimodal llm for real-time voice interactions,'' 2025.
\newblock Available at \url{https://github.com/fixie-ai/ultravox}.

\bibitem{shao2025}
Y.~J. Shao, H.~Zope, Y.~Jiang, {\em et~al.}, ``Future of work with ai agents: Auditing automation and augmentation potential across the u.s. workforce,'' {\em arXiv preprint arXiv:2506.06576}, 2025.

\end{thebibliography}

\end{document}